\documentclass[10pt,twocolumn,letterpaper]{article}
\pdfoutput=1

\usepackage{wacv}
\usepackage{times}
\usepackage{epsfig}
\usepackage{graphicx}
\usepackage{amsmath}
\usepackage{amssymb}
\usepackage{array}
\usepackage{stfloats}
\usepackage{float}

\usepackage[ruled,norelsize]{algorithm2e}
\usepackage{tabulary}

\makeatletter
\newcommand{\removelatexerror}{\let\@latex@error\@gobble}
\makeatother

\wacvfinalcopy 

\def\wacvPaperID{1296} 

\ifwacvfinal\fi
\begin{document}

\title{BOBBY2: Buffer Based Robust High-Speed Object Tracking}

\author{Keifer Lee \hspace{2cm} Jun Jet Tai \hspace{2cm} Swee King Phang\\
Taylor's University, 1 Jalan Taylor's, 47500, Subang Jaya, Selangor, Malaysia\\
{\tt\small leekeifer@sd.taylors.edu.my, junjet.tai02@sd.taylors.edu.my, sweeking.phang@taylors.edu.my}
}

\maketitle
\ifwacvfinal\thispagestyle{empty}\fi

\begin{abstract}
   In this work, a novel high-speed single object tracker that is robust against non-semantic distractor exemplars is introduced; dubbed BOBBY2. It incorporates a novel exemplar buffer module that sparsely caches the target's appearance across time, enabling it to adapt to potential target deformation. In addition, we demonstrate that exemplar buffer is capable of providing redundancies in case of unintended target drifts, a desirable trait in any middle to long term tracking. Even when the buffer is predominantly filled with distractors instead of valid exemplars, BOBBY2 is capable of maintaining a near-optimal level of accuracy. In terms of speed, BOBBY2 utilises a stripped down AlexNet as feature extractor with 63\% less parameters than a vanilla AlexNet, thus being able to run at 85 FPS. An augmented ImageNet-VID dataset was used for training with the one cycle policy, enabling it to reach convergence with less than 2 epoch worth of data. For validation, the model was benchmarked on the GOT-10k dataset and on an additional small, albeit challenging custom UAV dataset collected with the TU-3 UAV.
\end{abstract}


\section{Introduction}

Object tracking is one of the perennial tasks in the domain of computer vision and robotic perception amongst many other vision tasks due to the generality of its nature \cite{koller1993model,bradski1998real,burt1989object,stephens1990real,yilmaz2006object}.

A particular interests as of late is to perform single object tracking (SOT) tasks on unmanned aerial vehicles (UAVs), especially light-weight multi-rotor units, due to their wide-ranging applicability across industrial domains~\cite{bokeno2018package,peeters2015providing}. However, there are significant challenges facing such efforts in terms of computational efficiency, and robustness against confusing non-target instances (herein referred to as \textit{distractors}). The limited payload and battery capacity of a typical multi-rotor UAV restricts the type of allowable computing device on board to light-weight embedded systems, thereby severely constraining the computing power available. On the other hand, tracking from a mid-flight UAV usually entails having low target-to-image ratio amidst complex environment - \eg  low resolution scale and aspect ratio varying targets, drastic camera movement and occlusion - possibly littered with distractors. Thus, for feasible object tracking to be performed locally on a typical UAV, it would necessarily need to be both computationally efficient and robust in the aforementioned scenarios. This work is a step towards the direction realizing such applications by introducing a novel high-speed and robust SOT model.

Generally, the current methods on approaching object tracking can be segregated into the two categories of correlation filter based trackers and fully convolutional trackers. Siamese fully convolutional trackers that learn the task in an end-to-end manner have been shown by Betinetto \etal \cite{bertinetto2016fully} to be a simple yet highly competitive approach to hand-crafted object trackers, setting new state-of-the-art results in tracking capability. However, as with most deep learning models it is too computationally expensive for on-board embedded systems to run optimally as is. The same efficiency problem persists for subsequent follow-up Siamese convolutional trackers~\cite{zhu2018distractor,li2018high,bertinetto2016fully,valmadre2017end}. Thus, the work in locally performed SOT for UAVs have so far been relying on hand-crafted trackers and correlation filter models in order to meet the real-time demand on a budget~\cite{xue2018unmanned,wu2017vision,yang2019real,uzkent2018enkcf}. However, even with the more computationally efficient algorithms, both manually designed and learned, the computational overhead is still too high for optimal performance on a low-end embedded device. 
Correlation filters and other handcrafted trackers such as LSST~\cite{yang2019real} and CACFT~\cite{xue2018unmanned} achieves impressive accuracy and average run-time performance of 41 FPS and 19 FPS respectively. However, the latter correlation filter models are much more prone to the problem of ghosting and class confusion.

The challenges above form the basis of this study where a simple yet highly efficient Siamese convolutional SOT is introduced, dubbed BOBBY2. In order to decrease the computational overhead, BOBBY2 utilizes a combination of sparse feature extraction, lightweight feature extractors, and run-time heuristics to improve run-time FPS significantly compared to its closest counterpart, SiamFC~\cite{bertinetto2016fully}. As for robustness against distractors, a novel buffer base exemplar module is incorporated into the tracker to perform feature aggregation across time, and artificially generated negative samples were introduced in training to help the model learn better discriminative representations. Thus, BOBBY2 is capable of achieving 85 FPS with the full tracking pipeline, and 700 FPS as a standalone model. In addition, we have also demonstrated robustness of BOBBY2 against distractors by explicitly inserting confusing distractors as exemplars during tracking on the ImageNet-VID~\cite{ILSVRC15} dataset; it is capable of maintaining near-optimal performance even with a collection of distractor-majority exemplars in the buffer. For validation, the tracker is evaluated on the GOT-10k dataset~\cite{huang2018got} and a small but challenging custom UAV dataset collected with TU-3, a quad-rotor UAV on the campus parking lot. Another thing of note is the training regime used~\cite{smith2018disciplined,smith2018super}  - one cycle learning with super-convergence, enabling the model to be trained to convergence with just less than 2 epoch worth of data on ImageNet-VID.

The following sections are laid out in the following manner - exposition of related works in Section 2, discussion on the methodology employed in Section 3, review of the experimental results on accuracy and computational efficiency in Section 4, and a brief concluding discussion in Section 5.

\section{Related Works}

\subsection{Deep Object Tracker}
The predominant architecture used by current state-of-the-art single object trackers is of the Siamese CNN variant~\cite{zhu2018distractor,li2018high,bertinetto2016fully,valmadre2017end,wang2019fast}. In this design, two convolutional feature extractors, $\mathit{\varphi_1,\varphi_2}$ are used to ingest an image of a scene $\mathit{x}$ and the target $\mathit{z}$ (or exemplar) respectively, whereby the resulting feature maps are combined before being fed to a discriminative classifier $\mathit{g(\textbf{X},\textbf{Z})}$. In practice, it is common to use a single feature extractor to featurize both scene and exemplar in order to obtain a set of comparable feature maps. The intuition behind Siamese trackers could be construed as similarity learning, whereby the classifier tries to find the target patch from an approximately equivalent scene through a set of commonly computed feature maps $\mathit{f(x,z) = g(\varphi(x),\varphi(z))}$.

Though Siamese networks have already been introduced earlier in other visual tasks~\cite{taigman2014deepface,zagoruyko2015learning,koch2015siamese}, it was first applied to the domain of object tracking independently by Betinetto \etal \cite{bertinetto2016fully} with SiamFC and Held \etal \cite{held2016learning} with GOTURN, setting new state-of-the-art results, and have subsequently spawned other improved variants of the same architecture~\cite{zhu2018distractor,li2018high,valmadre2017end,wang2019fast}. Subsequent Siamese trackers have since then sought to improve in three general dimensions, accuracy (\eg EAO, AUC), robustness against distractors and run-time FPS. Models such as CFNet~\cite{valmadre2017end} and SiamRPN~\cite{li2018high} introduced correlation filters which are computationally cheaper to reduce the overall computation overhead. The former performs online update to the correlation filters during inference while the latter pose the problem as a one-shot learning task, computing the filter parameters from the target's initialization frame only, and performs region proposal during tracking. DaSiamRPN~\cite{zhu2018distractor} is an improvement of SiamRPN in terms of robustness against distractor instances by leveraging on a hard-negative mining scheme during training and custom loss terms to surgically penalize both inter-class and intra-class distractors. This enables DaSiamRPN to achieve significant advantages in long-term tracking scenarios at real-time \cite{kristan2018sixth}. Quantitatively, CFNet is capable of obtaining an average of 75 FPS with competitive accuracy on a NVIDIA Titan X GPU, 160 FPS for SiamRPN on a NVIDIA GTX1060 GPU, and an average of 135 FPS for DaSiamRPN on a powerful device with a NVIDIA Titan X GPU and 48GB RAM. Evidently, though Siamese trackers are simple and robust against distractors, they are still computationally unviable for resource constrained real-time tracking on UAVs. 

\begin{figure*}
\centering
    \includegraphics[width =\textwidth]{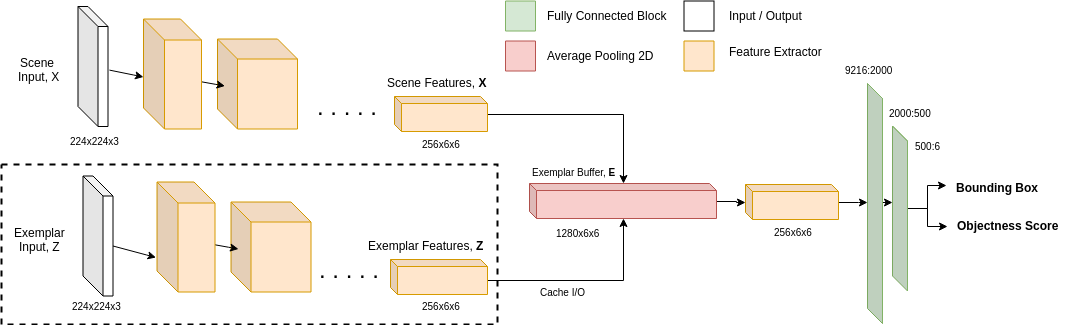}
    \caption{Overview of BOBBY2's architecture. The dashed box denotes the sparsely activate exemplar feature extractor; activated every $\mathit{\xi}$ frames. The feature extractor for both branches share the same underlying parameters for equivalent representation. The aggregated exemplar features are concatenated along the channels in the buffer (shown $\mathit{\beta}_{max} = 4$) before being pass to the classifier.}
    \label{fig:bobbyarc}
\end{figure*}

\subsection{Real-Time Single Object Tracking}

As one may surmise, applications such as real-time tracking on UAV thus far has largely avoided deep learning models and opted for the significantly more computationally efficient hand-crafted trackers, most of them with closed-form solutions \cite{xue2018unmanned,wu2017vision,yang2019real,uzkent2018enkcf}. Correlation filters in particular are a very popular in such time-sensitive use cases, though they lack the robustness of deep learning models \cite{bolme2010visual,henriques2014high,danelljan2014accurate}. Yang \etal's LSST tracker reformulates large least-square functions into smaller and more tractable forms in the Fourier domain to be solved using the efficient Recursive Least-Square algorithm (RLS) \cite{yang2019real}.Xue \etal \cite{xue2018unmanned}'s proposed CACFT, a SOT tracking framework utilizing correlation filters with fused feature maps and a conditional sparse exemplar template updating module to dynamically adapt to changing target properties. In \cite{wu2017vision}, Wu \etal proposed FOLT, a framework that utilizes Kalman Filters to estimate the target's state properties such as velocity, width and height with a fast discriminative saliency map generator for tracking. Uzkent and Seo \cite{uzkent2018enkcf} in turn extended the use of kernelized correlation filters (KCF) by ensembling, dubbed EnKCF for tracking. In their work, a series of KCFs are applied in succession across a series of frames mediated by a particle filter.

In terms of run-time performance, EnKCF obtains the highest FPS of 378 FPS on average, albeit at the cost of accuracy and robustness. FOLT followed with 141 FPS, LSST with 41 FPS and CACFT with 19 FPS. However, even these highly efficient algorithms may prove to be sub-optimal due to the devices used for the aforementioned benchmarks being still more powerful than a typical embedded system level device. FOLT was implemented on a device with an Intel Xeon W3250 CPU, 16GB RAM and in C++, LSST on an Intel i7-6700hq CPU with 4GB RAM, CACFT on a desktop class Intel i5-7500 CPU with 16GB RAM, and EnKCF on an unspecified desktop class platform. Though these figures are not directly comparable due to hardware and software differences, it could still offer useful insights into their relative performances. Another disadvantage of these hand-crafted models are their higher susceptibility to the problem of ghosting and class confusion compared to deep models.

\section{Methodology and Procedures}

\subsection{Overview of BOBBY2}
BOBBY2 is an end-to-end learned and convolutional Siamese object tracker with an exemplar buffer for feature aggregation. Feature aggregation as defined in the current work is the accumulation of features derived from the target across time in order to form a series of spatio-temporally representative exemplars. Figure~\ref{fig:bobbyarc} is a simplified illustration of the network architecture behind BOBBY2.

A pre-trained AlexNet \cite{krizhevsky2012imagenet} is used as the backbone feature extractor to produce a rich representation for the input scene and exemplars. AlexNet is used instead of more advance architectures such as VGG~\cite{simonyan2014very} and ResNet~\cite{he2016deep} because the former has better computational efficiency compared to the latter models. In addition, though AlexNet is a shallower model, the resulting features are representative enough for the task of template matching based object tracking. Common mobile-oriented model networks such as SqueezeNet has been empirically found to be less optimal in terms of actual run-time FPS. Despite having lower parameter count and storage size, these deeper networks have higher memory demand, which is a significant bottleneck for run-time FPS performance.

One of the main novelty of BOBBY2 is the feature buffer module that caches sparse key exemplar frames across time. Inputs of scene, $\mathit{x}_i$ are ingested by the scene feature extractor at every frame and the exemplar, $\mathit{z}_i$ only at frames delineated by the hyperparameter, $\mathit{\xi}$. During in-between frames that do not correspond to the $\mathit{\xi}$ series, the tracker behaves as an asymmetrical Siamese network by deactivating the exemplar branch. Formally defined as 

\begin{equation} \label{high_fp}
f(x_i,\textbf{E}_i) = g(\varphi(x_i),\varphi(\textbf{E}_i))
\end{equation} 

\noindent where $\mathit{g}$ is the template matching and target verification function, $\mathit{\varphi}$ is the feature extractor, and $\textbf{E}$ the exemplar buffer which in turn is defined as

\begin{equation} \label{exem_buffer}
\textbf{E} = \{\textbf{Z}_1,\textbf{Z}_2,\textbf{Z}_3,\cdots,\textbf{Z}_{\beta_{max}}\}
\end{equation} 

\noindent where $\mathit{\beta_{max}}$ is the upper-bound size of the exemplar buffer. The exemplar feature instances of the aggregated buffer are interrelated in time with respect to $\mathit{\xi}$:

\begin{equation} \label{ex_int}
frame_{Z_{j+1}} = frame_{Z_{j}} + \xi
\end{equation} 

Thus, the exemplar feature extractor is sparsely activated at specific intervals which endows higher computational efficiency as the interval grows. Ideally, the interval $\mathit{\xi}$ is varied depending on the complexity of a given sequence. Sequences with higher degree and rate of frame-to-frame target variation or confusing instances necessitates a shorter activation interval and vice versa, though that may not always be possible. As for the buffer size, $\mathit{\beta}_{max}$ of 4 was empirically found to be a suitable value with acceptable trade-offs in terms of robustness and efficiency.

\subsection{Scene-Objectness Classification}

BOBBY2 poses tracking as a multitask problem by jointly learning to determine the bounding box coordinates of the target for localization, and a scene-objectness score to verify the target's presence in the scene itself:

\begin{equation} \label{Loss FX1}
L_{joint} = y_{obj}L_{bbox} + \alpha L_{obj}
\end{equation} 

\noindent where $\mathit{y}_{obj}$ is the label for scene-objectness and $\alpha$ the scene-objectness loss multiplier. 

For the bounding box loss, the pytorch Smooth-L1 loss function was used. Thus, if $\mathit{|x_{s,m}-y_{s,m}| <} ~1$, Equation~\ref{Loss FX2} was used as the bounding box loss function, otherwise Equation~\ref{Loss FX3}.

\begin{equation} \label{Loss FX2}
L_{bbox} = \frac{1}{8M} \sum_{s=1}^{4} \sum_{m=1}^{M} (x_{s,m} - y_{s,m})^2
\end{equation} 

\begin{equation}\label{Loss FX3}
L_{bbox} = \frac{1}{4M} \sum_{s=1}^{4} \sum_{m=1}^{M} (|x_{s,m} - y_{s,m}| - 0.5)
\end{equation}

\noindent where $\mathit{s}$ are the sides of the bounding box, and $\mathit{M}$ the length of the dataset. The scene-objectness loss is simply the Cross-Entropy Loss with 2 classes corresponding to target absence and presence in a scene. The bounding box loss is only enforced if a target is indeed present in a scene (scene-objectness label of 1). Otherwise, the network was only penalize with the amplified classification loss. 

In order to do that,  an additional 378,606 negative samples were generated from the ImageNet-VID datasets \cite{ILSVRC15}, constituting approximately 30\% of the total samples\footnote{The exact collection of negative samples and the general negatives generating scripts can be found in the project Github repository.}. This is done to encourage the network to learn of more discriminative features beyond plain template matching, and to hedge against false positive instances where the target is absent. The combination of a scene-objectness score and an exemplar buffer enables BOBBY2 to offer advantages over trackers on the two different ends of the exemplar template updating spectrum; tracking by greedy exemplar update at every frame, and by initialization-only exemplar. Unlike the former costly class of greedy trackers that are susceptible to problems such as potential target drifts from bad predictions, BOBBY2 is much more robust in that regard by having a series of curated exemplar for redundancies. Yet, unlike the latter class of trackers that utilizes only the initialization frame as a template, BOBBY2 is capable of adapting to significant target deformation and variance across time with its sparse exemplar updates.

\begin{figure}
    \begin{center}
    \includegraphics[width=\columnwidth]{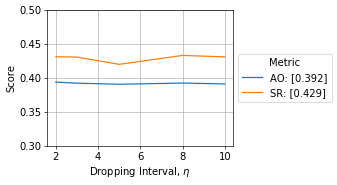}
    \caption{Average Overlap (AO) and Success Rate (SR) with respect to varying $\mathit{\eta}$ on the GOT-10k dataset. Demonstrated minimal performance drop when $\mathit{\eta = 2}$.}
    \label{fig:dropping_range}
    \end{center}
\end{figure}

\subsection{Approaching Real-Time Performance}
BOBBY2 employs a lightweight feature extractor and a heuristical trick of frame-dropping in order to achieve real-time performance. As compared to the vanilla AlexNet, our implementation is much more lightweight, having less than half the parameters of a vanilla AlexNet despite having 4 times more exemplar information cached in the buffer. Shown in Table~\ref{table:FE}.

\begin{table}
\begin{center}
\begin{tabular}{ccc}
\hline
\textbf{Model}	& \textbf{Parameter Count}	& \textbf{FLOPS} \\
\hline
AlexNet & 61.10M & 4.29G  \\
BOBBY2 & \textbf{22.23M} & \textbf{4.13G} \\
\hline
\end{tabular}
\end{center}
\caption{Computation overhead comparison.}
\label{table:FE}
\end{table}

In addition, the every frames coinciding with the dropping interval, $\mathit{\eta}$ will be dropped in time constrained scenarios in order to fulfill the required response time. We show in Figure~\ref{fig:dropping_range} that even for the most extreme case of $\mathit{\eta = 2}$ where every interleaving frames are dropped,our model is robust enough to maintain its optimal tracking accuracy and precision, as shown in Figure~\ref{fig:dropping_range}. The approximate speed-up factor (SF) with respect to the dropping interval is given by the following simple equation.

\begin{equation}\label{Speedup}
SF = \frac{\eta}{\eta - 1}
\end{equation}

\subsection{Training and Evaluation}

\begin{figure}
    \begin{center}
    \includegraphics[width=\columnwidth]{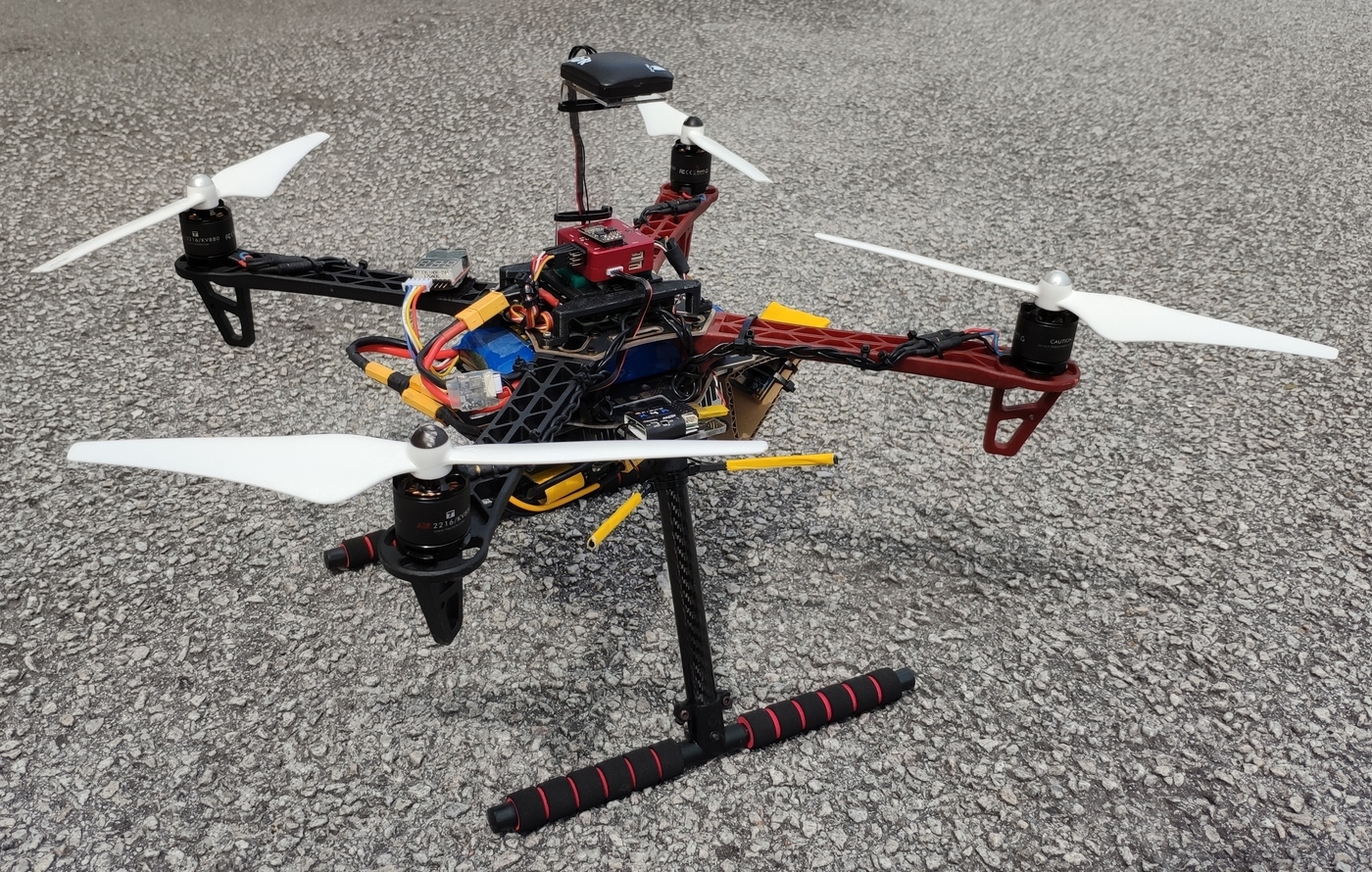}
    \caption{The TU-3 UAV, constructed from a DJI F450 platform with an ELP webcamera and a Raspberry Pi module.}
    \label{fig:drone}
    \end{center}
\end{figure}

\begin{figure}[ht]
\centering
    \includegraphics[width=\columnwidth]{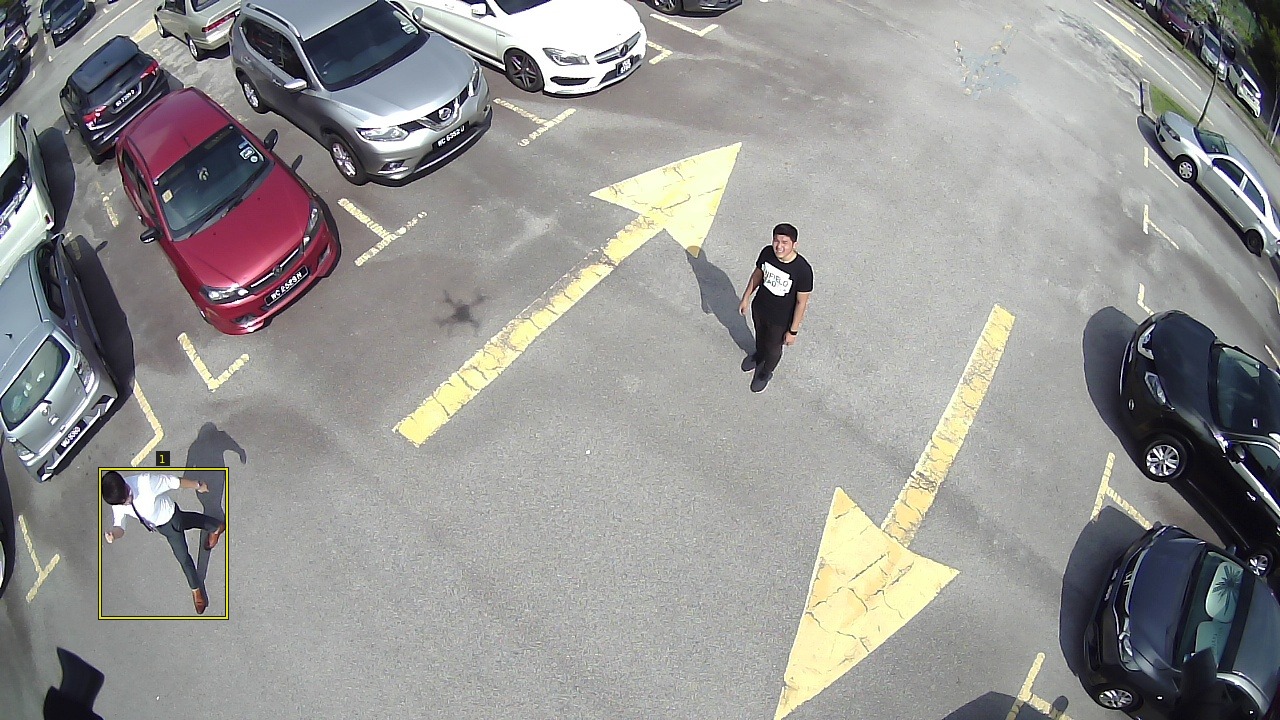}\\
    \includegraphics[width=\columnwidth]{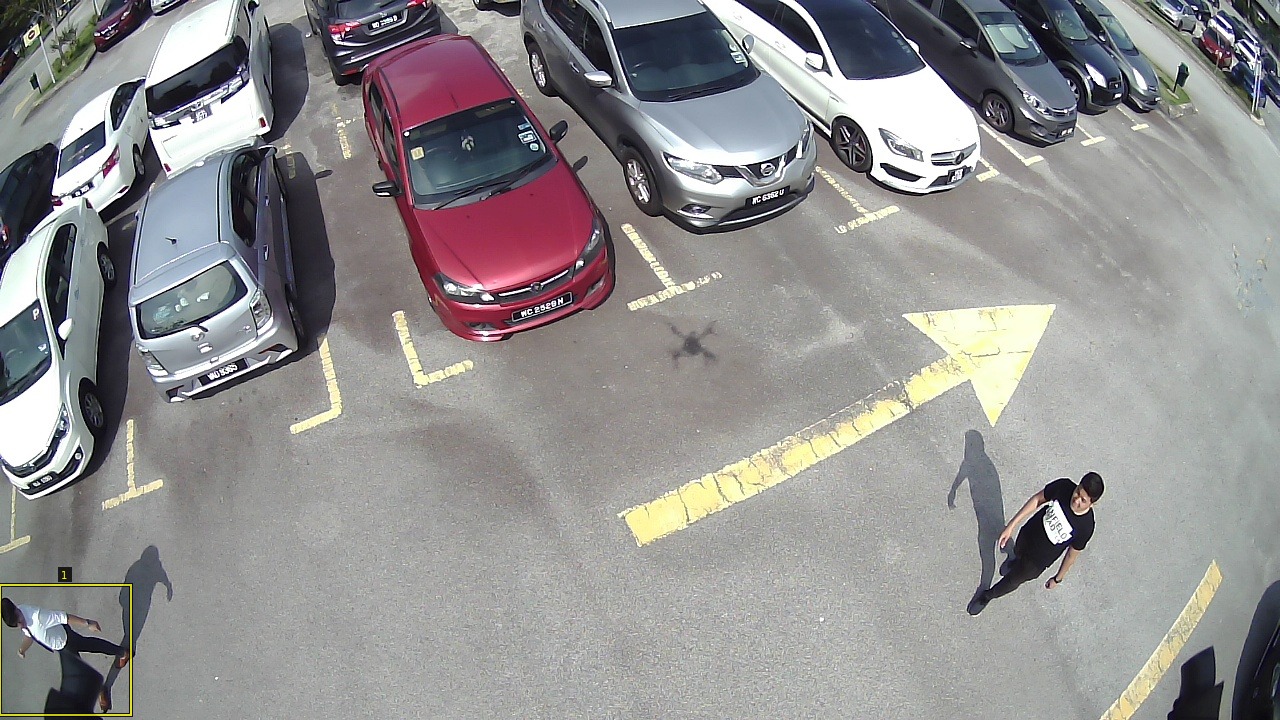}\\
    \includegraphics[width=\columnwidth]{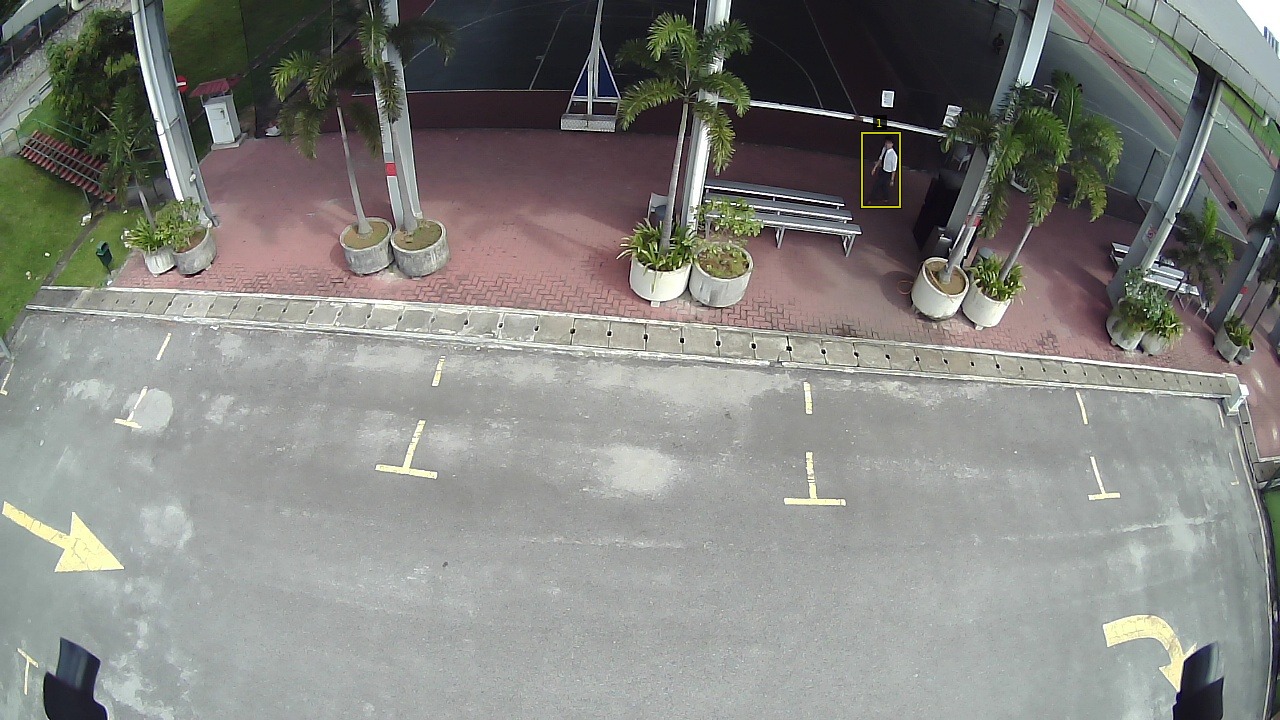}
    \caption{Sample of frames from the custom dataset from TU-3 UAV. Target is the man in white framed by a yellow bounding box.}\label{fig:TU3_dataset}
\end{figure}

The model was trained with positive and artificial negatives sample from the ImageNet-VID dataset~\cite{ILSVRC15}. For training, the Adam optimizer~\cite{kingma2014adam} was used with in conjunction with a cyclical learning rate in accordance with the one cycle training policy~\cite{smith2018disciplined}. With that, we were able to train BOBBY2 in just 5 cycles, each comprising of a small random subset - 25\% or 352,879 samples - of the training dataset by leveraging on the super-convergence effect introduced by Smith and Topin~\cite{smith2018super}. The weight-decay value used was 10e-2, a scene-objectness loss multiplier $\alpha$ of 10, batch size of 32 and learning rate of 3e-3.\footnote{Plots of losses can be found in Appendix A.}

In terms of evaluation, BOBBY2 was benchmarked on the GOT10-K, and ImageNet-VID dataset. For further validation, we have collected a small sample of footages taken from a custom UAV platform codenamed TU-3. It is a UAV platform constructed by our UAV research group, with a standard DJI F450 platform, customized with an ELP webcamera connected to a Raspberry Pi for image logging (See Figure~\ref{fig:drone}). The footages are highly challenging with characteristics such as target disappearance, full occlusions, camera warp, target deformation and presence of confusing instances, shown in Figure~\ref{fig:TU3_dataset}.

\section{Experiments and Result Evaluations}

\begin{figure}
    \begin{center}
    \includegraphics[width = \columnwidth]{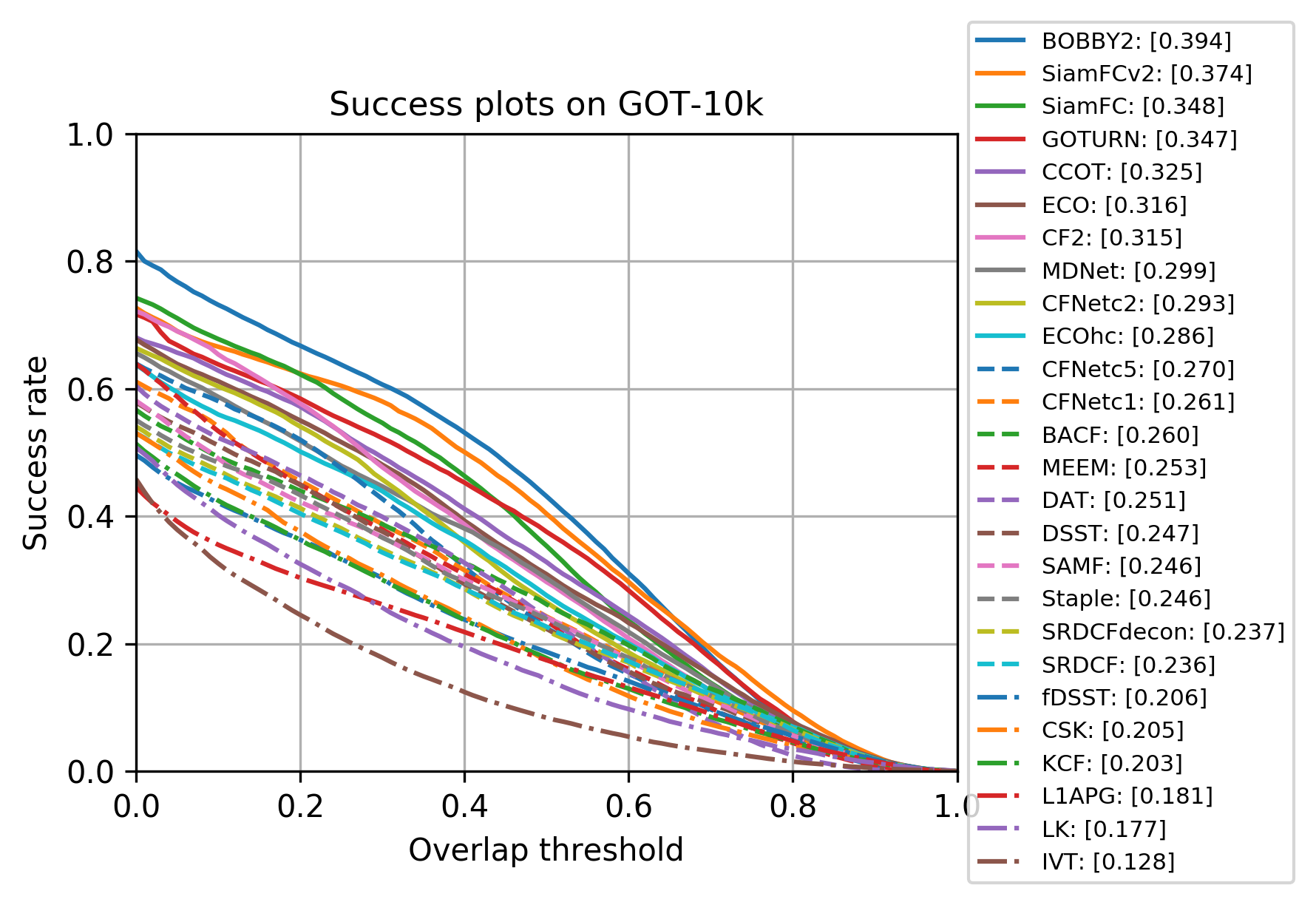}
    \caption{Success plot on the GOT-10k dataset.}
    \label{fig:got_success}
    \end{center}
\end{figure}

\subsection{Accuracy and Precision}

In this section, analysis of the corresponding benchmarks are presented. First, results from the GOT-10k benchmark is presented to provide a comparison of performance in terms of Average Overlap (AO). Thereafter, we turn to a closer analysis on the tracker's performance in different configurations and scenarios to highlight its robustness, and conclude with a brief review of its performance on the custom UAV dataset to demonstrate BOBBY2's generalization capability.

\begin{figure*}
    \begin{center}
    \includegraphics[width=\textwidth]{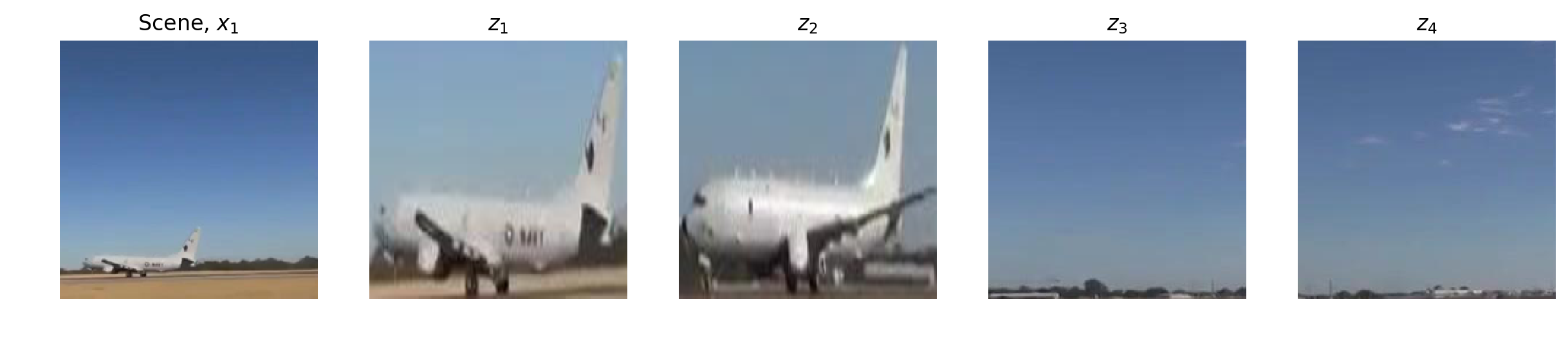}
    \caption{Visualization of the input stack with 50\% distractor in the exemplar buffer. $z_1$ and $z_2$ are positive exemplars while $z_3$ and $z_4$ are distractors.}
    \label{fig:posneg}
    \end{center}
\end{figure*}

\paragraph{Performance on GOT-10k:} To evaluate the performance, we had use the metrics as put forth by the dataset authors, Average Overlap (AO) and Success Rate (SR). The former is the mean overlap of the predicted bounding box and the ground truth, while the latter is the fraction of frames that have an overlap value of higher than 0.5. Figure~\ref{fig:got_success} is the success plot of a collection of prominent SOT ranked by their area under the curve (AUC) scores. The top trackers all utilizes the convolution operation one way or another as expected. The top 3 performing trackers comprises of the SiamFC and SiamFC2 trackers with a score of 0.374 and 0.348 respectively. BOBBY2 outperforms all other trackers in this category, with $5.3\%$ improvement in AUC over SiamFC22 and $13.2\%$ over SiamFC. Other non-convolutional and non-deep models such as KCF performed significantly worst than those that are. GOTURN, one of the first convolutional Siamese trackers, performed better than later hybrid correlation-filter models such as CFNet. It has to be noted that unlike the trackers on the list, our model was not pre-trained on the GOT-10k training set as it was only trained on ImageNet-VID, demonstrating its transferability across dataset. As for FPS performance, a direct comparison could not be drawn due to the difference in hardware used in the benchmark of each tracker. Therefore the following discussion on FPS serves only to provide a rough comparison for completeness. The fastest tracker by raw FPS is the Circulant Structure Kernel (CSK)~\cite{henriques2012exploiting} tracker with 122 FPS, followed by GOTURN with 109 FPS, and the Kernelized Correlation Filter (KCF) tracker with 88 FPS. Both CSK and KCF were benchmarked on a on a 16-core CPU only, while GOTURN uses an NVIDIA Titan X GPU. BOBBY2 comes in with 85 FPS on an NVIDIA 2070-Super GPU, significantly faster than its closest counterpart, SiamFC which poses 24 FPS with an NVIDIA Titan X GPU. The raw speed of BOBBY2 without the tracking pipeline is capable of running at an average of 700 FPS on the same platform. Upon closer inspection, the bottleneck in the tracking pipeline was found to be the image cropping-resizing procedure, which accounts for 93\% of the total run-time.

\paragraph{Performance on ImageNet-VID:} One particular feature that makes BOBBY2 stand out is its exemplar buffer. This is of importance in tracking with models that performs dynamic update of the exemplar which allows it to adapt to the target's deformations across time, crucial for long term tracking. However, this comes at a cost of increased chances for tracking failure on mid to long sequences, whereby any errors in the exemplar update will rapidly accumulate and cascade across frames. Another downside of such adaptive scheme is the increased computation overhead from needing to perform additional feature extraction at certain interval - the exemplar buffer refresh interval of BOBBY2 is dictated by $\xi$. We show that our model is able to perform exactly such adaptive updates while being sufficiently robust against distractors in its buffer, hence negating the problem of error accumulation significantly. In addition, our model imposes no additional computational overhead from performing feature extraction for the mid-sequence exemplars updates by dropping the frames after featurizing during such instances without incurring significant penalty to accuracy, as demonstrated previously in Figure~\ref{fig:dropping_range}. This way, we are able to maintain a consistent computation cost of approximately a single forward pass (assuming no additional frame dropping is performed, $\mathit{\eta=\infty}$) at each iteration throughout the tracking process. 

We test the aforementioned distractor robustness by performing controlled tracking runs over a range of distractor percentages whereby the exemplar buffer is explicitly filled with non-semantic distractors instead of a valid exemplar. Figure~\ref{fig:posneg} is a visualization of the tracker's scene input and exemplar buffer state for a single sample during said test. Table~\ref{table:posneg} is a detailed tabulation of its performance. We have conducted the test with a distractor percentage of 0\% to 75\%, and have found that model holds up very well with only a 1\% drop in success rate at 0.5 AO, even when the majority of the buffer contents are only distractors. Therefore, it has been demonstrated that the exemplar buffer acts as redundancy in such cases, capable of hedging against target drifts due to error accumulation during adaptive tracking.

\begin{table*}
\begin{center}
\begin{tabular}[width=\textwidth]{ccccc}
\hline
\textbf{\% Distractor}	& \textbf{SR @ 0.25 AO}	& \textbf{SR @ 0.50 AO} & \textbf{SR @ 0.75 AO} & \textbf{SR @ 0.90 AO}  \\
\hline
0\% & 0.922 & 0.845 & 0.663 & 0.363 \\
25\% & 0.924 & 0.851 & 0.665 & 0.357 \\
50\% & 0.920 & 0.844 & 0.649 & 0.350 \\
75\% & 0.916 & 0.835 & 0.619 & 0.335 \\
\hline
\end{tabular}
\end{center}
\caption{Average Success Rate of BOBBY2 across different percentage of buffer exemplar distractors.}
\label{table:posneg}
\end{table*}

\paragraph{Performance on TU-3 UAV:} For additional validation, the model was evaluated on a small yet challenging custom UAV dataset. Furthermore, no fine-tuning was performed on this or any other additional dataset to further demonstrate the transferability and generalization capability of our model. Table~\ref{table:hornet_ana} is a breakdown of the results. On this particular dataset, the tracker had difficulty in providing optimal results, and the drop in SR as the percentage of buffer distractor increases is much more noticeable than seen in Table~\ref{table:posneg}. The SR drop between having 0\% and 75\% buffer distractor for the AO threshold of 0.25 in the latter is approximately 0.65\% and 20\% in the current dataset. We attribute this to the significant difference in footage characteristics between the training ImageNet-VID dataset and the custom UAV dataset. We believe that fine-tuning the tracker on these new challenging scenes will offer non-trivial improvement. It should be noted however that despite not being trained at all on said dataset, BOBBY2 still exhibited an enduring robustness in its performance even with a distractor-majority buffer.

\begin{table*}
\begin{center}
\begin{tabular}{cccc}
\hline
\textbf{\% Distractor}	& \textbf{SR @ 0.25 AO}	& \textbf{SR @ 0.50 AO} & \textbf{SR @ 0.75 AO}  \\
\hline
0\% & 0.474 & 0.111 & 0.010  \\
25\% & 0.456 & 0.101 & 0.005 \\
50\% & 0.444 & 0.087 & 0.003 \\
75\% & 0.395 & 0.063 & 0.003 \\
\hline
\end{tabular}
\end{center}
\caption{Average Success Rate of BOBBY2 across different percentage of buffer exemplar distractors on custom UAV dataset.}
\label{table:hornet_ana}
\end{table*}

\section{Conclusion}

In this work, a novel buffer based Siamese object tracker is introduced. Specifically in our tests, a buffer size $\beta = 4$ was used. Hence, at each forward pass, the tracker has access to 4 times more exemplar data than a conventional Siamese tracker. Yet, it has 63\% less parameters than a vanilla AlexNet and a slightly lower number of floating point operations. On the GOT-10k dataset, BOBBY2 manages to outperform all the trackers being benchmarked, with 5.3\% and 13.2\% AUC improvements over the next two highest performing trackers. There gap between BOBBY2 and non-deep learning trackers are even more significant, as expected. In terms of speed, our tracker is capable of running above real-time requirements at 85 FPS. One of the heuristic trick used for computation speed improvement is frame-dropping. In that regard, we have also shown that the tracker is capable of maintaining a near-optimal accuracy even with an extreme frame-dropping regime where every interleaved frames are dropped, $\eta = 2$. As for robustness, BOBBY2 has a significant advantage over other adaptive trackers in dealing with the problem of potential target drifts due to error accumulation in exemplar updates. We have shown that it could once again maintain a near-optimal accuracy in such scenarios, even when 75\% of the exemplars in the buffer are non-semantic distractors. For a final validation, BOBBY2 was evaluated on a challenging custom UAV dataset collected with the TU-3 UAV, without fine-tuning. The performance on said dataset is sub-optimal - SR of 0.474 at 0.25 AO without exemplar distractors - due to its high difficulty, and the significant difference between it and the training dataset. Despite that, the performance is sufficient to demonstrate the generalizability of BOBBY2 across data of different distributions; this can be non-trivially improved with additional fine-tuning.

The current work deals only with the domain of SOT with minimal long term tracking capability. Thus, we look to extend BOBBY2 to account for Multi Object Tracking and Long Term Object tracking cases in our future work. We conjecture that the exemplar buffer will offer a non-trivial advantage in these domains as well due to its adaptive yet robust buffered embeddings. Another concern is to improve the run-time FPS by overcoming the severe cropping-resizing bottleneck in the tracking pipeline. In addition, we seek to subject future versions of BOBBY2 to model compression, weight pruning and quantization techniques in order to further improve on the speed performance.

{\small
\bibliographystyle{ieee}
\bibliography{biblio}
}

\appendix
\section{Appendices}

\subsection{Appendix A}\label{app_loss}
\begin{figure}[H]
    \begin{center}
    \includegraphics[width=0.8\columnwidth]{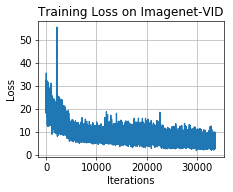}
    \caption{Training loss of BOBBY2 on the augmented ImageNet-VID dataset.}
    \label{fig:train}
    \end{center}
\end{figure}

\begin{figure}[H]
    \begin{center}
    \includegraphics[width=0.8\columnwidth]{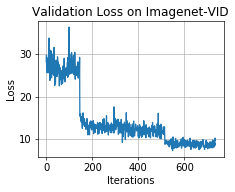}
    \caption{Validation loss of BOBBY2 on the augmented ImageNet-VID dataset.}
    \label{fig:validation}
    \end{center}
\end{figure}

\end{document}